\documentclass[11pt]{article}
\usepackage{coling2016}
\usepackage{graphicx}
\usepackage{dsfont} 
\usepackage{algorithm}
\usepackage{algorithmic}
\usepackage{multirow}
\usepackage{epsfig}
\usepackage{graphicx}
\usepackage{epstopdf}
\usepackage{times}
\usepackage{url}
\usepackage{latexsym}
\usepackage{helvet}
\usepackage{courier}
\usepackage{amsmath,amssymb,amsfonts}
\usepackage{bigstrut}
\usepackage{MnSymbol}
\usepackage{array}
\usepackage{lscape,longtable}
\usepackage{bigstrut}
\usepackage{adjustbox}
\usepackage{tikz-dependency}
\usepackage{subcaption}

\title{A Deeper Look into Sarcastic Tweets\\Using Deep Convolutional Neural Networks}

\author{Soujanya Poria, Erik Cambria, Devamanyu Hazarika, Prateek Vij\\
	Nanyang Technological University\\
	 50 Nanyang Ave, Singapore 639798\\
	{\tt \{sporia, cambria\}@ntu.edu.sg}\\
	{\tt\{devamanyu, prateek\}@sentic.net}	
	}

\date{}

\begin{document}
\maketitle
\begin{abstract}
Sarcasm detection is a key task for many natural language processing tasks. In sentiment analysis, for example, sarcasm can flip the polarity of an ``apparently positive'' sentence and, hence, negatively affect polarity detection performance. 
To date, most approaches to sarcasm detection have treated the task primarily as a text categorization problem. Sarcasm, however, can be expressed in very subtle ways and requires a deeper understanding of natural language that standard text categorization techniques cannot grasp. In this work, we develop models based on a pre-trained convolutional neural network for extracting sentiment, emotion and personality features for sarcasm detection. Such features, along with the network's baseline features, allow the proposed models to outperform the state of the art on benchmark datasets. We also address the often ignored generalizability issue of classifying data that have not been seen by the models at learning phase.
\end{abstract}

\blfootnote{
	\hspace{-0.65cm}  
This work is licensed under a Creative Commons Attribution 4.0 International Licence. Licence details: http://creativecommons.org/licenses/by/4.0/}

\section{Introduction}
Sarcasm is defined as ``a sharp, bitter, or cutting expression or remark; a bitter gibe or taunt''.
As the fields of affective computing and sentiment analysis have gained increasing popularity \cite{camacsa}, it is a major concern to detect sarcastic, ironic, and metaphoric expressions. Sarcasm, especially, is key for sentiment analysis as it can completely flip the polarity of opinions.
Understanding the ground truth, or the facts about a given event, allows for the detection of contradiction between the objective polarity of the event (usually negative) and its sarcastic characteristic by the author (usually positive), as in ``\textit{I love the pain of breakup}''. Obtaining such knowledge is, however, very difficult. 

In our experiments, we exposed the classifier to such knowledge extracted indirectly from Twitter. Namely, we used Twitter data crawled in a time period, which likely contain both the sarcastic and non-sarcastic accounts of an event or similar events. We believe that unambiguous non-sarcastic sentences provided the classifier with the ground-truth polarity of those events, which the classifier could then contrast with the opposite estimations in sarcastic sentences. Twitter is a more suitable resource for this purpose than blog posts, because the polarity of short tweets is easier to detect (as all the information necessary to detect polarity is likely to be contained in the same sentence) and because the Twitter API makes it easy to collect a large corpus of tweets containing both sarcastic and non-sarcastic examples of the same event.

Sometimes, however, just knowing the ground truth or simple facts on the topic is not enough, as the text may refer to other events in order to express sarcasm. For example, the sentence ``\textit{If Hillary wins, she will surely be pleased to recall Monica each time she enters the Oval Office :P :D}'', which refers to the 2016 US presidential election campaign and to the events of early 1990's related to the US president Clinton, is sarcastic because Hillary, a candidate and Clinton's wife, would in fact not be pleased to recall her husband's alleged past affair with Monica Lewinsky. The system, however, would need a considerable amount of facts, commonsense knowledge, anaphora resolution, and logical reasoning to draw such a conclusion. In this paper, we will not deal with such complex cases.

Existing works on sarcasm detection have mainly focused on unigrams and the use of emoticons~\cite{gonzalez2011identifying,carvalho2009clues,barbieri2014modelling}, unsupervised pattern mining approach~\cite{maynard2014cares}, semi-supervised approach~\cite{riloff2013sarcasm} and n-grams based approach~\cite{tsur2010icwsm,davidov2010semi,ptacek2014sarcasm,joshi2015harnessing} with sentiment features. Instead, we propose a framework that learns sarcasm features automatically from a sarcasm corpus using a convolutional neural network (CNN). We also investigate whether features extracted using the pre-trained sentiment, emotion and personality models can improve sarcasm detection performance. Our approach uses relatively lower dimensional feature vectors and outperforms the state of the art on different datasets. In summary, the main contributions of this paper are the following:
\begin{itemize}
	\item To the best of our knowledge, this is the first work on using deep learning for sarcasm detection.
	\item Unlike other works, we exploit sentiment and emotion features for sarcasm detection. As user profiling is also an important factor for detecting sarcastic content, moreover, we use personality-based features for the first time in the literature.
	\item Pre-trained models are commonly used in computer vision. In the context of natural language processing (NLP), however, they are barely used. Hence, the use of pre-trained models for feature extraction is also a major contribution of this work.
\end{itemize}

The rest of the paper is organized as follows: Section~\ref{sec:rel} proposes a brief literature review on sarcasm detection; Section~\ref{sec:approach} presents the proposed approach; experimental results and thorough discussion on the experiments are given in Section~\ref{sec:exp}; finally, Section~\ref{sec:conclusion} concludes the paper.

\section{Related Works}
\label{sec:rel}
NLP research is gradually evolving from lexical to compositional semantics \cite{camjum} through the adoption of novel meaning-preserving and context-aware paradigms such as convolutional networks~\cite{porasp}, recurrent belief networks \cite{chadee}, statistical learning theory \cite{onesta}, convolutional multiple kernel learning \cite{porcon}, and commonsense reasoning~\cite{camsen}. But while other NLP tasks have been extensively investigated, sarcasm detection is a relatively new research topic which has gained increasing interest only recently, partly thanks to the rise of social media analytics and sentiment analysis.
 Sentiment analysis \cite{zadeh2016multimodal} and using multimodal information as a new trend  \cite{zadeh2016mosi,poria2015deep,traffickingacl17,soujanyaacl17,porcon} is a popular branch of NLP research that aims to understand sentiment of documents automatically using combination of various machine learning approaches \cite{zadeh2015micro,tensoremnlp17,soujanyaacl17,ceclm17}. 

An early work in this field was done by \cite{tsur2010icwsm} on a dataset of 6,600 manually annotated Amazon reviews using a kNN-classifier over punctuation-based and pattern-based features, i.e., ordered sequence of high frequency words.
\cite{gonzalez2011identifying} used support vector machine (SVM) and logistic regression over a feature set of unigrams, dictionary-based lexical features and pragmatic features (e.g., emoticons) and compared the performance of the classifier with that of humans.
\cite{reymul} described a set of textual features for recognizing irony at a linguistic level, especially in short texts created via Twitter, and constructed a new model that was assessed along two dimensions: representativeness and relevance.
\cite{riloff2013sarcasm} used the presence of a positive sentiment in close proximity of a negative situation phrase as a feature for sarcasm detection. 
\cite{liebrecht2013perfect} used the Balanced Window algorithm for classifying Dutch tweets as sarcastic vs.~non-sarcastic; n-grams (uni, bi and tri) and intensifiers were used as features for classification.

\cite{buschmeier2014impact} compared the performance of different classifiers on the Amazon review dataset using the imbalance between the sentiment expressed by the review and the user-given star rating. 
Features based on frequency (gap between rare and common words), written spoken gap (in terms of difference between usage), synonyms (based on the difference in frequency of synonyms) and ambiguity (number of words with many synonyms) were used by \cite{barbieri2014modelling} for sarcasm detection in tweets.
\cite{joshi2015harnessing} proposed the use of implicit incongruity and explicit incongruity based features along with lexical and pragmatic features, such as emoticons and punctuation marks. Their method is very much similar to the method proposed by \cite{riloff2013sarcasm} except \cite{joshi2015harnessing} used explicit incongruity features. Their method outperforms the approach by \cite{riloff2013sarcasm} on two datasets.

\cite{ptacek2014sarcasm} compared the performance with different language-independent features and pre-processing techniques for classifying text as sarcastic and non-sarcastic. The comparison was done over three Twitter dataset in two different languages, two of these in English with a balanced and an imbalanced distribution and the third one in Czech. The feature set included n-grams, word-shape patterns, pointedness and punctuation-based features.

In this work, we use features extracted from a deep CNN for sarcasm detection. Some of the key differences between the proposed approach and existing methods include the use of a relatively smaller feature set, automatic feature extraction, the use of deep networks, and the adoption of pre-trained NLP models.

\section{Sentiment Analysis and Sarcasm Detection}
Sarcasm detection is an important subtask of sentiment analysis \cite{cambria2015clsa}. Since sarcastic sentences are subjective, they carry sentiment and emotion-bearing information. Most of the studies in the literature~\cite{joshi2016automatic,bosdev,joshi2015harnessing,fariro} include sentiment features in sarcasm detection with the use of a state-of-the-art sentiment lexicon. Below, we explain how sentiment information is key to express sarcastic opinions and the approach we undertake to exploit such information for sarcasm detection.

\begin{figure}[h]
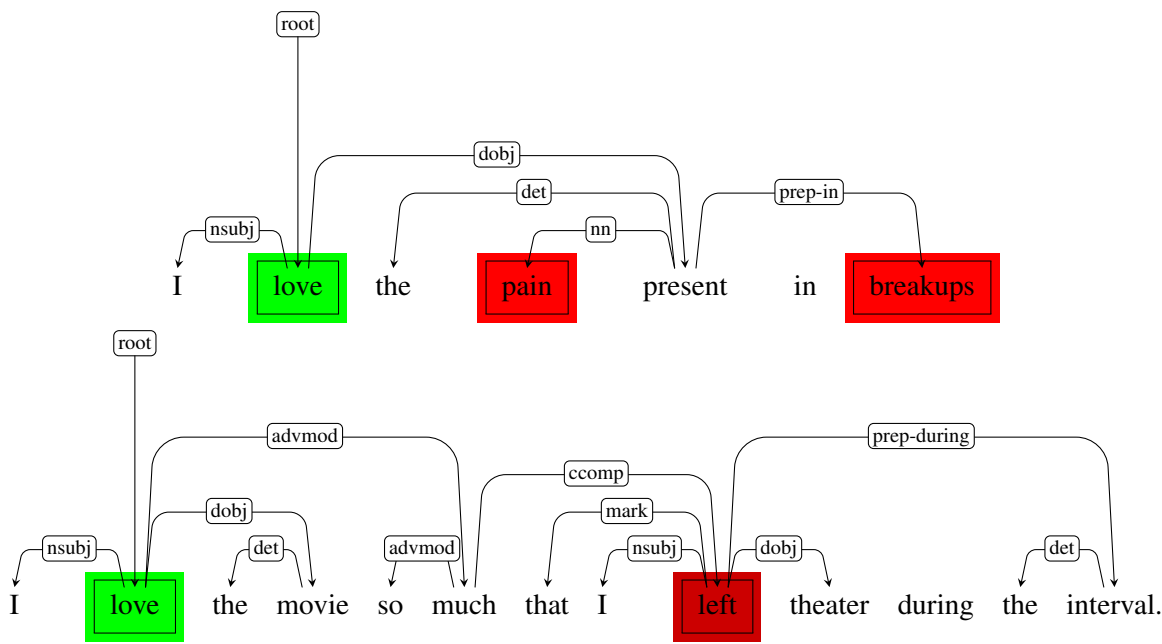

 \begin{center}
 \begin{dependency}
 	\begin{deptext}[column sep =0.2cm]
 		I \&[.5cm]  \colorbox[rgb]{0,1,0}{\framebox{~love~\strut}} 
 		 \& the \&[.5cm] 
 		\colorbox[rgb]{1,0,0}{\framebox{~pain~\strut}} 
 		\&[.5cm] 
 	 present
 		\&[0.5cm]
in
\&
 		\colorbox[rgb]{1,0,0}{\framebox{~breakups~\strut}} 
\\
 	\end{deptext}
 	\depedge{2}{1}{nsubj}
 	\depedge{5}{3}{det}
 	\depedge{5}{4}{nn}
 	\depedge{2}{5}{dobj}
 	\deproot[edge unit distance=5ex]{2}{root}
 	\depedge{5}{7}{prep-in}
 \end{dependency}
 \end{center}
   \begin{center}
  	\begin{dependency}
  		\begin{deptext}[column sep =0.2cm]
  			I \&[.5cm]  \colorbox[rgb]{0,1,0}{\framebox{~love~\strut}} 
  			\& the \& 
			movie
  			\&
  			so
  			\&
  			much
  			\&
		that
		\&
		I
		\&[.5cm]
		\colorbox[rgb]{0.8,0,0}{\framebox{~left~\strut}} 
		\&
		theater
		\&
		during
		\&
		the
		\&
		interval.
  			\\
  		\end{deptext}
  		\depedge{2}{1}{nsubj}
  		\depedge{4}{3}{det}
  		\depedge{2}{4}{dobj}
  		\depedge{6}{5}{advmod}
  		\depedge{2}{6}{advmod}
  		\deproot[edge unit distance=5ex]{2}{root}
  		\depedge{9}{7}{mark}
  		\depedge{9}{8}{nsubj}
  		\depedge{6}{9}{ccomp}
  		\depedge{9}{10}{dobj}
  		\depedge{13}{12}{det}
  		\depedge{9}{13}{prep-during}
  	\end{dependency}
  \end{center}
  \caption{Sentiment shifting can be an indicator of sarcasm.}
    \label{fig:sentiment}
\end{figure}

In general, most sarcastic sentences contradict the fact. In the sentence ``I love the pain present in the breakups" (Figure \ref{fig:sentiment}), for example, the word ``love" contradicts ``pain present in the breakups'', because in general no-one loves to be in pain. In this case, the fact (i.e., ``pain in the breakups") and the contradictory statement to that fact (i.e., ``I love") express sentiment explicitly. Sentiment shifts from positive to negative but, according to sentic patterns~\cite{poria2015sentiment}, the literal sentiment remains positive. Sentic patterns, in fact, aim to detect the polarity expressed by the speaker; thus, whenever the construction ``I love'' is encountered, the sentence is positive no matter what comes after it (e.g., ``I love the movie that you hate''). In this case, however, the sentence carries sarcasm and, hence, reflects the negative sentiment of the speaker.

In another example (Figure \ref{fig:sentiment}), the fact, i.e., ``I left the theater during the interval", has implicit negative sentiment. The statement ``I love the movie" contradicts the fact ``I left the theater during the interval"; thus, the sentence is sarcastic. Also in this case the sentiment shifts from positive to negative and hints at the sarcastic nature of the opinion.
 
The above discussion has made clear that sentiment (and, in particular, sentiment shifts) can largely help to detect sarcasm. In order to include sentiment shifting into the proposed framework, we train a sentiment model for sentiment-specific feature extraction. Training with a CNN helps to combine the local features in the lower layers into global features in the higher layers. We do not make use of sentic patterns~\cite{poria2015sentiment} in this paper but we do plan to explore that research direction as a part of our future work.
In the literature, it is found that sarcasm is user-specific too, i.e., some users have a particular tendency to post more sarcastic tweets than others. This acts as a primary intuition for us to extract personality-based features for sarcasm detection. 

\section{The Proposed Framework}
\label{sec:approach}
As discussed in the literature \cite{riloff2013sarcasm}, sarcasm detection may depend on sentiment and other cognitive aspects. For this reason, we incorporate both sentiment and emotion clues in our framework. Along with these, we also argue that personality of the opinion holder is an important factor for sarcasm detection. In order to address all of these variables, we create different models for each of them, namely: sentiment, emotion and personality. The idea is to train each model on its corresponding benchmark dataset and, hence, use such pre-trained models together to extract sarcasm-related features from the sarcasm datasets. 

\begin{figure}[h]
	\centering
	\includegraphics[scale=0.57]{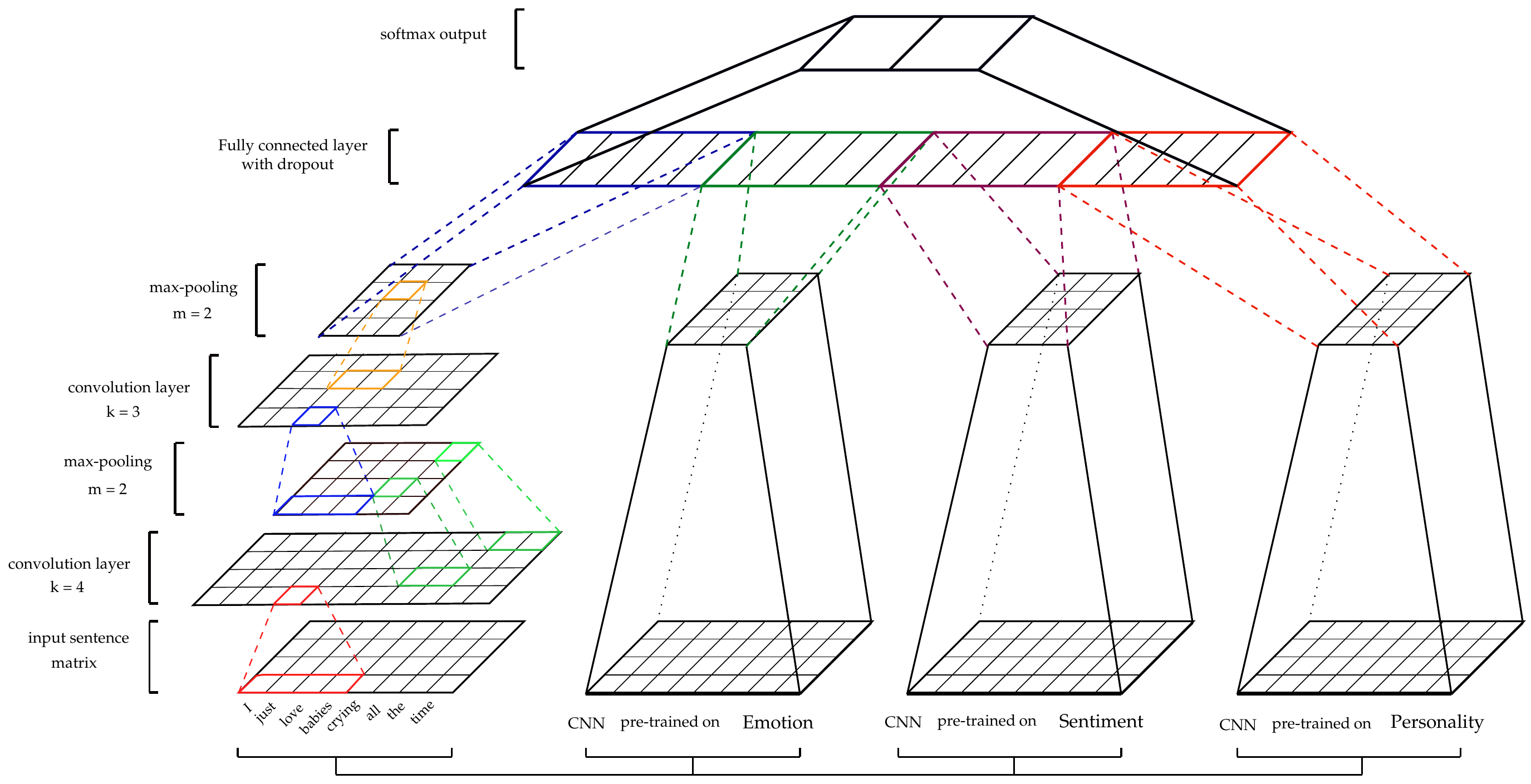}
	\caption{The proposed framework: deep CNNs are combined together to detect sarcastic tweets.}
	\label{fig:framework}
\end{figure}

Now, the viable research question here is - Do these models help to improve sarcasm detection performance?'
Literature shows that they improve the performance but not significantly. Thus, do we need to consider those factors in spotting sarcastic sentences? Aren't n-grams enough for sarcasm detection? Throughout the rest of this paper, we address these questions in detail.
The training of each model is done using a CNN. Below, we explain the framework in detail. Then, we discuss the pre-trained models. Figure \ref{fig:framework} presents a visualization of the proposed framework.

\subsection{General CNN Framework}
\label{sec:CNN}
CNN can automatically extract key features from the training data. It grasps contextual local features from a sentence and, after several convolution operations, it forms a global feature vector out of those local features. CNN does not need the hand-crafted features used in traditional supervised classifiers. 
Such hand-crafted features are difficult to compute and a good guess for encoding the features is always necessary in order to get satisfactory results. CNN, instead, uses a hierarchy of local features which are important to learn context. The hand-crafted features often ignore such a hierarchy of local features. 

Features extracted by CNN can therefore be used instead of hand-crafted features, as they carry more useful information.
The idea behind convolution is to take the dot product of a vector of $k$ weights $w_{k}$ also known as kernel vector with each $k$-gram in the sentence $s(t)$ to obtain another sequence of features $c(t)=(c_{1}(t),c_{2}(t),\ldots,c_{\mathrm{L}}(t))$. 
\begin{eqnarray}
{c}_{j}={{w}_{k}}^{T}.{\boldsymbol{\mathrm{x}}_{i:i+k-1}}
\label{eqn:conv1}
\end{eqnarray}

Thus, a max pooling operation is applied over the feature map and the maximum value $\hat{c}(t)=\mathrm{max}\{\boldsymbol{\mathrm{c}}(t)\}$ is taken as the feature corresponding to this particular kernel vector. Similarly, varying kernel vectors and window sizes are used to obtain multiple features \cite{RSM:Nal2014}. 
For each word $x_{i}$ in the vocabulary, a $d$-dimensional vector representation is given in a look up table that is learned from the data \cite{mikolov2013efficient}. The vector representation of a sentence, hence, is a concatenation of vectors for individual words. Similarly, we can have look up tables for other features. One might want to provide features other than words if these features are suspected to be helpful. The convolution kernels are then applied to word vectors instead of individual words. 

We use these features to train higher layers of the CNN, in order to represent bigger groups of words in sentences. We denote the feature learned at hidden neuron $h$ in layer $l$ as $F^{l}_{h}$. Multiple features may be learned in parallel in the same CNN layer. The features learned in each layer are used to train the next layer:
\begin{eqnarray}
F^{l}={\sum}_{h=1}^{n_{h}}w_{k}^{h}*F^{l-1}
\end{eqnarray}

where * indicates convolution and $w_{k}$ is a weight kernel for hidden neuron $h$ and $n_{h}$ is the total number of hidden neurons.
The CNN sentence model preserves the order of words by adopting convolution kernels of gradually increasing sizes that span an increasing number of words and ultimately the entire sentence. 
As mentioned above, each word in a sentence is represented using word embeddings. 

\paragraph{\bf Word Embeddings}
We employ the publicly available word2vec
vectors, which were trained on 100 billion words from Google
News. The vectors are of dimensionality 300, trained using the continuous bag-of-words
architecture \cite{mikolov2013efficient}. Words not present in the set of pre-trained words are initialized randomly. 
However, while training the neural network, we use non-static representations. These include the word vectors, taken as input, into the list of parameters to be learned during training. 

Two primary reasons motivated us to use non-static channels as opposed to static ones. Firstly, the common presence of informal language and words in tweets resulted in a relatively high random initialization of word vectors due to the unavailability of these words in the \textit{word2vec} dictionary. Secondly, sarcastic sentences are known to include polarity shifts in sentimental and emotional degrees. For example, ``\textit{I love the pain present in breakups}" is a sarcastic sentence with a significant change in sentimental polarity. As \textit{word2vec} was not trained to incorporate these nuances, we allow our models to update the embeddings during training in order to include them.
Each sentence is wrapped to a window of $n$, where $n$ is the maximum number of words amongst all sentences in the dataset. We use the output of the fully-connected layer of the network as our feature vector. 

\paragraph{CNN-SVM}
We have done two kinds of experiments: firstly, we used CNN for the classification; secondly, we extracted features from the fully-connected layer of the CNN and fed them to an SVM for the final classification. The latter CNN-SVM scheme is quite useful for text classification as shown by Poria et al.~\cite{poria2015deep}. We carry out n-fold cross-validation on the dataset using CNN. In every fold iteration, in order to obtain the training and test features, the output of the fully-connected layer is treated as features to be used for the final classification using SVM.
Table \ref{table:set} shows the training settings for each CNN model developed in this work. \emph{ReLU} is used as the non-linear activation function of the network\footnote{We show the optimal training settings of the CNNs used in this work. Changing kernels' size or adding/removing layers does not improve results.}. 
The network configurations of all models developed in this work are given in Table \ref{table:set}.

		\begin{table}
			\centering
		\begin{tabular}{|ccccccccc|}
			\hline
			 &
			\multicolumn{2}{c}{Convolution Layer 1} &
			1st Max &
			\multicolumn{2}{c}{Convolution Layer 2} &
			2nd Max-&
			FC &
			Softmax\\
			\cline{2-3}\cline{5-6}
			&\small Kernel Size&\small Feature Map&Pooling&\small Kernel Size&\small Feature Map&Pooling&Layer&Output\\
			\hline
			S&4,5&50&2&3&100&2&100&3\\
			E&3,4,5&50&2&2&100&2&150&6\\
			P&3,4,5&50&2&2&100&2&150&2\\	
			B &4,5&50&2&3&100&2	&100&2\\
			\hline			
		\end{tabular}
		\caption{Training settings for each deep model. Legenda: FC = Fully-Connected, S = Sentiment model, E = Emotion model, P = Personality model, B = Baseline model.}
		\label{table:set}
		\end{table}

\subsection{Sentiment Feature Extraction Model}
As discussed above, sentiment clues play an important role for sarcastic sentence detection. In our work, we train a CNN (see Section~\ref{sec:CNN} for details) on a sentiment benchmark dataset. This pre-trained model is then used to extract features from the sarcastic datasets. In particular, we use Semeval 2014 \cite{rosenthal2014semeval} Twitter Sentiment Analysis Dataset for the training. This dataset contains 9,497 tweets out of which 5,895 are positive, 3,131 are negative and 471 are neutral.
The fully-connected layer of the CNN used for sentiment feature extraction has 100 neurons, so 100 features are extracted from this pre-trained model. The final softmax determines whether a sentence is positive, negative or neutral. Thus, we have three neurons in the softmax layer.
\subsection{Emotion Feature Extraction Model}
We use the CNN structure as described in Section~\ref{sec:CNN} for emotional feature extraction. 
As a dataset for extracting emotion-related features, we use the corpus developed by \cite{aman2007identifying}. This dataset consists of blog posts labeled by their corresponding emotion categories. As emotion taxonomy, the authors used six basic emotions, i.e., Anger, Disgust, Surprise, Sadness, Joy and Fear. In particular, the blog posts were split into sentences and each sentence was labeled. The dataset contains 5,205 sentences labeled by one of the emotion labels.
After employing this model on the sarcasm dataset, we obtained a 150-dimensional feature vector from the fully-connected layer. As the aim of training is to classify each sentence into one of the six emotion classes, we used six neurons in the softmax layer.

\subsection{Personality Feature Extraction Model}
Detecting personality from text is a well-known challenging problem. In our work, we use five personality traits described by \cite{matthews1999personality}, i.e., Openness, Conscientiousness, Extraversion, Agreeableness, and Neuroticism, sometimes abbreviated as OCEAN (by their first letters). 
As a training dataset, we use the corpus developed by \cite{matthews1999personality}, which contains 2,400 essays labeled by one of the five personality traits each.

The fully-connected layer has 150 neurons, which are treated as the features. We concatenate the feature vector of each personality dimension in order to create the final feature vector. Thus, the personality model ultimately extracts a 750-dimensional feature vector (150-dimensional feature vector for each of the five personality traits). This network is replicated five times, one for each personality trait. In particular, we create a CNN for each personality trait and the aim of each CNN is to classify a sentence into binary classes, i.e., whether it expresses a personality trait or not.

\subsection{Baseline Method and Features}
\label{sec:baseline}
CNN can also be employed on the sarcasm datasets in order to identify sarcastic and non-sarcastic tweets. 
We term the features extracted from this network \emph{baseline features}, the method as \emph{baseline method} and the CNN architecture used in this baseline method as \emph{baseline CNN}. Since the fully-connected layer has 100 neurons, we have 100 baseline features in our experiment. This method is termed baseline method as it directly aims to classify a sentence as sarcastic vs non-sarcastic. The baseline CNN extracts the inherent semantics from the sarcastic corpus by employing deep domain understanding. The process of using baseline features with other features extracted from the pre-trained model is described in Section~\ref{sec:merge}.

\begin{figure}[h]
	\centering
	\includegraphics[scale=0.19]{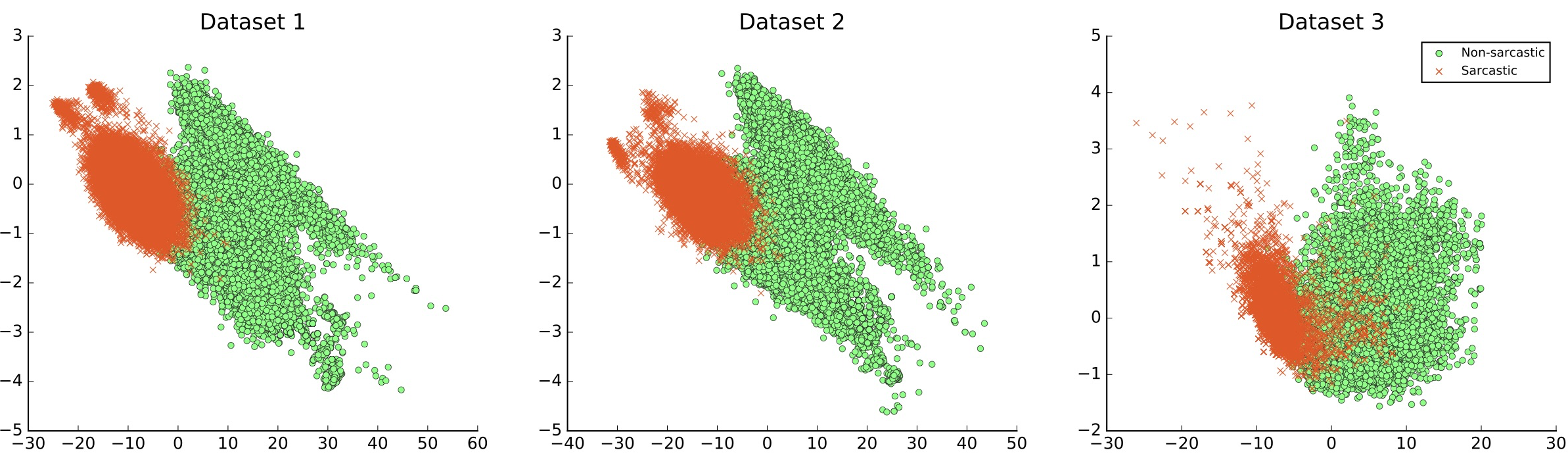}
	\caption{Visualization of the data.}
	\label{fig:dataset}
\end{figure}

\section{Experimental Results and Discussion}
\label{sec:exp}
In this section, we present the experimental results using different feature combinations and compare them with the state of the art. For each feature we show the results using only CNN and using CNN-SVM (i.e., when the features extracted by CNN are fed to the SVM). Macro-F1 measure is used as an evaluation scheme in the experiments.
\subsection{Sarcasm Datasets Used in the Experiment}
\label{sec:datasets}
\paragraph{Dataset 1 (Balanced Dataset)}
This dataset was created by \cite{ptacek2014sarcasm}. The tweets were downloaded from Twitter using \#sarcasm as a marker for sarcastic tweets. It is a monolingual English dataset which consists of a balanced distribution of 50,000 sarcastic tweets and 50,000 non-sarcastic tweets.

\paragraph{Dataset 2 (Imbalanced Dataset)}
Since sarcastic tweets are less frequently used~\cite{ptacek2014sarcasm}, we also need to investigate the robustness of the selected features and the model trained on these features on an imbalanced dataset. To this end, we used another English dataset from \cite{ptacek2014sarcasm}. It consists of 25,000 sarcastic tweets and 75,000 non-sarcastic tweets.

\paragraph{Dataset 3 (Test Dataset)}
We have obtained this dataset from The Sarcasm Detector\footnote{http://thesarcasmdetector.com}. It contains 120,000 tweets, out of which 20,000 are sarcastic and 100,000 are non-sarcastic. We randomly sampled 10,000 sarcastic and 20,000 non-sarcastic tweets from the dataset. Visualization of both the original and subset data show similar characteristics.

\paragraph{Pre-processing}
\label{sec:preprocess}
A two-step methodology has been employed in filtering the datasets used in our experiments. Firstly, we identified and removed all the ``user", ``URL" and ``hashtag" references present in the tweets using efficient regular expressions. Special emphasis was given to this step to avoid traces of hashtags, which might trigger the models to provide biased results.
Secondly, we used \emph{NLTK Twitter Tokenizer} to ensure proper tokenization of words along with special symbols and emoticons. Since our deep CNNs extract contextual information present in tweets, we include emoticons as part of the vocabulary. This enables the emoticons to hold a place in the word embedding space and aid in providing information about the emotions present in the sentence.

\subsection{Merging the Features}
\label{sec:merge}
Throughout this research, we have carried out several experiments with various feature combinations. For the sake of clarity, we explain below how the features extracted using difference models are merged.

\begin{itemize}
	\item In the standard feature merging process, we first extract the features from all deep CNN based feature extraction models and then we concatenate them. Afterwards, SVM is employed on the resulted feature vector.
	\item In another setting, we use the features extracted from the pre-trained models as the static channels of features in the CNN of the baseline method. These features are appended to the hidden layer of the \emph{baseline CNN}, preceding the final output softmax layer.
\end{itemize}

For comparison, we have re-implemented the state-of-the-art methods. Since \cite{joshi2015harnessing} did not mention about the sentiment lexicon they use in the experiment, we used SenticNet \cite{camnt4} in the re-implementation of their method.

\subsection{Results on Dataset 1}
As shown in Table \ref{table:exp}, for every feature CNN-SVM outperforms the performance of the CNN. 
Following \cite{tsur2010icwsm}, we have carried out a 5-fold cross-validation on this dataset.
The baseline features (\ref{sec:baseline}) perform best among other features. Among all the pre-trained models, the sentiment model (F1-score: 87.00\%) achieves better performance in comparison with the other two pre-trained models. Interestingly, when we merge the baseline features with the features extracted by the pre-trained deep NLP models, we only get 0.11\% improvement over the F-score. It means that the baseline features alone are quite capable to detect sarcasm. On the other hand, when we combine sentiment, emotion and personality features, we obtain 90.70\% F1-score. This indicates that the pre-trained features are indeed useful for sarcasm detection. 
We also compare our approach with the best research study conducted on this dataset (Table \ref{table:comparison}). Both the proposed baseline model and the \emph{baseline + sentiment + emotion + personality} model outperform the state of the art \cite{joshi2015harnessing,ptacek2014sarcasm}. 
One important difference with the state of the art is that \cite{ptacek2014sarcasm} used relatively larger feature vector size ($>$500,000) than we used in our experiment (1,100). This not only prevents our model to overfit the data but also speeds up the computation. Thus, we obtain an improvement in the overall performance with automatic feature extraction using a relatively lower dimensional feature space.

In the literature, word n-grams, skipgrams and character n-grams are used as baseline features. According to Ptacek et al.~\cite{ptacek2014sarcasm}, these baseline features along with the other features (sentiment features and part-of-speech based features) produced the best performance. However, Ptacek et al.~did not analyze the performance of these features when they were not used with the baseline features.
Pre-trained word embeddings play an important role in the performance of the classifier because, when we use randomly generated embeddings, performance falls down to 86.23\% using all features.

\subsection{Results on Dataset 2}
5-fold cross-validation has been carried out on Dataset 2. Also for this dataset, we get the best accuracy when we use all features. Baseline features have performed significantly better (F1-score: 92.32\%) than all other features. Supporting the observations we have made from the experiments on Dataset 1, we see CNN-SVM outperforming CNN on Dataset 2. However, when we use all the features, CNN alone (F1-score: 89.73\%) does not outperform the state of the art \cite{ptacek2014sarcasm} (F1-score: 92.37\%). As shown in Table \ref{table:comparison}, CNN-SVM on the \emph{baseline + sentiment + emotion + personality} feature set outperforms the state of the art (F1-score: 94.80\%). Among the pre-trained models, the sentiment model performs best (F1-score: 87.00\%). 

\begin{table}[h]
	\centering
	\small
	\begin{tabular}{|cccccccccc|}
		\hline
		\multirow{2}{*}{B} &
		\multirow{2}{*}{S} &
		\multirow{2}{*}{E} &
		\multirow{2}{*}{P} &
		\multicolumn{2}{c}{Dataset 1} &
		\multicolumn{2}{c}{Dataset 2} &
		\multicolumn{2}{c|}{Dataset 3} \\
		\cline{5-10}
		& & & & CNN & CNN-SVM & CNN & CNN-SVM & CNN & CNN-SVM \\
		\hline
		+&&&& 95.04\% &97.60\%&89.33\% &92.32\% &88.00\% &92.20\%\\
		&+&&& - &87.00\%& - &86.50\% &- &73.50\%\\
		&&+&& - &76.30\%&- &84.71\% &- &72.10\%\\
		&&&+& - &75.00\%&- &77.90\% &- &74.41\%\\
		&+&+&+&-&{\bf 90.70\%}&-&{\bf 90.90\%}&-&{\bf 84.43\%}\\
		+&+&&&95.21\%&97.67\%&89.69\%&94.60\%&88.58\%&93.12\%\\
		+&&+&&95.22\%&97.65\%&89.72\%&94.50\%&88.56\%&92.63\%\\
		+&&&+&95.21\%&97.64\%&89.62\%&93.60\%&88.26\%&92.50\%\\
		+&+&+&+&95.30\%&{\bf 97.71\%}&89.73\%&{\bf 94.80\%}&88.51\%&{\bf 93.30\%}\\
		\hline
	\end{tabular}
	\caption{Experimental Results. Legenda: B = Baseline, S = Sentiment, E = Emotion, P = Personality, 5-fold cross-validation is carried out for all the experiments.}
	\label{table:exp}
\end{table}

	\begin{table}
		\centering
			\small
		\begin{tabular}{|l cccc|} 
			\hline
	Method	&	Dataset 1 & Dataset 2 & Dataset 3 & D3 =$>$ D1\\ [0.5ex] 
	\hline
			\cite{ptacek2014sarcasm} &94.66\% &92.37\% &63.37\% &53.02\%\\	[0.5ex]
			\cite{joshi2015harnessing} &65.05\% &62.37\% &60.80\% &47.32\%\\[0.5ex]
			Proposed Method (using all features) &97.71\% &94.80\% &	93.30\%&76.78\%\\	[0.5ex]
			\hline
		\end{tabular}
		\caption{Performance comparison of the proposed method and the state-of-the-art approaches. Legenda: D3 =$>$ D1 is the model trained on Dataset 3 and tested on Dataset 1.}
		\label{table:comparison}
	\end{table}

Table \ref{table:exp} shows the performance of different feature combinations.
The gap between the F1-scores of only baseline features and all features is larger on the imbalanced dataset than the balanced dataset. This supports our claim that sentiment, emotion and personality features are very useful for sarcasm detection, thanks to the pre-trained models. The F1-score using sentiment features when combined with baseline features is 94.60\%. On both of the datasets, emotion and sentiment features perform better than the personality features. Interestingly, using only sentiment, emotion and personality features, we achieve 90.90\% F1-score.
	
\subsection{Results on Dataset 3}
Experimental results on Dataset 3 show the similar trends (Table \ref{table:comparison}) as compared to Dataset 1 and Dataset 2. The highest performance (F1-score 93.30\%) is obtained when we combine baseline features with sentiment, emotion and personality features. In this case, also CNN-SVM consistently performs better than CNN for every feature combination. The sentiment model is found to be the best pre-trained model. F1-score of 84.43\% is obtained when we merge sentiment, emotion and personality features.

Dataset 3 is more complex and non-linear 
in nature compared to the other two datasets. As shown in Table \ref{table:comparison}, the methods by \cite{joshi2015harnessing} and \cite{ptacek2014sarcasm} perform poorly on this dataset. The TP rate achieved by \cite{joshi2015harnessing} is only 10.07\% and that means their method suffers badly on complex data\footnote{We use \emph{RBF} kernel, C=8 and gamma=0.01 to evaluate the method of Joshi et al.~on Dataset 3 with 5-fold cross-validation.}. The approach of \cite{ptacek2014sarcasm} has also failed to perform well on Dataset 3, achieving 62.37\% with a better TP rate of 22.15\% than \cite{joshi2015harnessing}. On the other hand, our proposed model performs consistently well on this dataset achieving 93.30\%.

\subsection{Testing Generalizability of the Models and Discussions} 
To test the generalization capability of the proposed approach, we perform training on Dataset 1 and test on Dataset 3. The F1-score drops down dramatically to 33.05\%. In order to understand this finding, we visualize each dataset using PCA (Figure \ref{fig:dataset}). It depicts 
that, although Dataset 1 is mostly linearly separable, Dataset 3 is not. A linear kernel that performs well on Dataset 1 fails to provide good performance on Dataset 3. If we use \emph{RBF} kernel, it overfits the data and produces worse results than what we get using linear kernel. Similar trends are seen in the performance of other two state-of-the-art approaches \cite{joshi2015harnessing,ptacek2014sarcasm}. Thus, we decide to perform training on Dataset 3 and test on the Dataset 1. As expected better performance is obtained\footnote{We report the result using all the features in this case.} with F1-score 76.78\%. However, the other two state-of-the-art approaches fail to perform well in this setting. While the method by \cite{joshi2015harnessing} obtains F1-score of 47.32\%, the approach by \cite{ptacek2014sarcasm} achieves 53.02\% F1-score when trained on Dataset 3 and tested on Dataset 1. Below, we discuss about this generalizability issue of the models developed or referred in this paper.

As discussed in the introduction, sarcasm is very much topic-dependent and highly contextual. For example, let us consider the tweet ``I am so glad to see Tanzania played very well, I can now sleep well :P". Unless one knows that Tanzania actually did not play well in that game, it is not possible to spot the sarcastic nature of this sentence. Thus, an n-gram based sarcasm detector trained at time $t_{i}$ may perform poorly to detect sarcasm in the tweets crawled at time $t_{j}$ (given that there is a considerable gap between these time stamps) because of the diversity of the topics (new events occur, new topics are discussed) of the tweets. Sentiment and other contextual clues can help to spot the sarcastic nature in this kind of tweets. A highly positive statement which ends with a emoticon expressing joke can be sarcastic. 

State-of-the-art methods lack these contextual information which, in our case, we extract using pre-trained sentiment, emotion and personality models. Not only these pre-trained models, the baseline method (baseline CNN architecture) performs better than the state-of-the-art models in this generalizability test setting. In our generalizability test, when the pre-trained features are used with baseline features, we get 4.19\% F1-score improvement over the baseline features. On the other hand, when they are not used with the baseline features, together they produce 64.25\% F1-score.

Another important fact is that an n-grams model cannot perform well on unseen data unless it is trained on a very large corpus. If most of the n-grams extracted from the unseen data are not in the vocabulary of the already trained n-grams model, in fact, the model will produce a very sparse feature vector representation of the dataset. Instead, we use the word2vec embeddings as the source of the features, 
as word2vec allows for the computation of similarities between unseen data and training data.

\subsection{Baseline Features vs Pre-trained Features}
\begin{figure*}
	\centering
	\begin{subfigure}[t]{.5\textwidth}
		\centering
		\includegraphics[width=.7\linewidth]{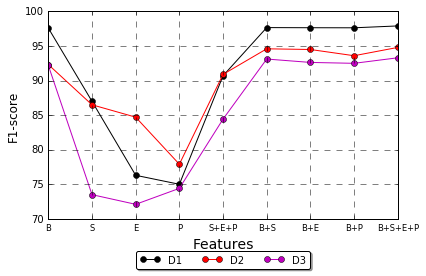}
		\caption{F1-score using different feature combinations.}
		\label{fig:sub1}
	\end{subfigure}%
	~
	\begin{subfigure}[t]{.5\textwidth}
		\centering
		\includegraphics[width=.7\linewidth]{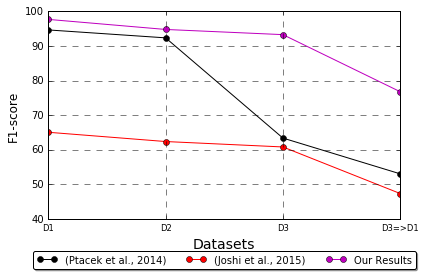}
		\caption{Comparison with the state of the art on benchmark datasets.}
		\label{fig:sub2}
	\end{subfigure}
	\caption{Plot of the performance of different feature combinations and methods.}
	\label{fig:plot}
\end{figure*}

Our experimental results show that the baseline features outperform the pre-trained features for sarcasm detection. However, the combination of pre-trained features and baseline features beats both of themselves alone. It is counterintuitive, since experimental results prove that both of those features learn almost the same global and contextual features. In particular, baseline network dominates over pre-trained network as the former learns most of the features learned by the latter. Nonetheless, the combination of baseline and pre-trained classifiers improves the overall performance and generalizability, hence proving their effectiveness in sarcasm detection.
Experimental results show that sentiment and emotion features are the most useful features, besides baseline features (Figure \ref{fig:plot}). Therefore, in order to reach a better understanding of the relation
between personality features among themselves and with other pre-trained features, we carried out Spearman correlation testing. Results, displayed in Table \ref{table:correlation}, show that those features are highly correlated with each other. 

	\begin{table}[h]
			\newcommand\Y{\hphantom{$^1$}}
			\newcommand\X{\%\Y}
		\centering
		\small
		\begin{tabular}{|l ccccc|} 
			\hline
			& Sentiment	& Happy	& Fear & Openness & Conscientiousness\\ [0.5ex] 
			Sentiment & 1.0 & 0.04$^*$ & 0.03$^*$ & 0.59$^*$ & 0.83$^*$\\
			Happy & & 1.0 & -0.48$^*$ & 0.14$^*$ & 0.12$^*$\\
			Fear & & & 1.0 & -0.10$^*$ & -0.09$^*$\\
			Openness & & & & 1.0 & 0.23$^*$\\
			Conscientiousness & & & & & 1.0\\
			\hline
		\end{tabular}
		\caption{Spearman's correlations between different features. * Correlation is significant at the 0.05 level.}
		\label{table:correlation}
	\end{table}
	
\section{Conclusion}
\label{sec:conclusion}
In this work, we developed pre-trained sentiment, emotion and personality models for identifying sarcastic text using CNN, which are found to be very effective for sarcasm detection. 
In the future, we plan to evaluate the performance of the proposed method on a large corpus and other domain-dependent corpora. Future work will also focus on analyzing past tweets and activities of users in order to better understand their personality and profile and, hence, further improve the disambiguation between sarcastic and non-sarcastic text.
\bibliography{clean_bib}
\bibliographystyle{acl}
\end{document}